\documentclass[table,xcdraw]{article}

\PassOptionsToPackage{numbers, compress}{natbib}

\usepackage[preprint]{neurips}

\usepackage[utf8]{inputenc} 
\usepackage[T1]{fontenc}    
\usepackage{hyperref}       
\usepackage{url}            
\usepackage{booktabs}       
\usepackage{amsfonts}       
\usepackage{nicefrac}       
\usepackage{microtype}      
\usepackage{amsmath}
\usepackage{graphicx}
\usepackage{xcolor}
\definecolor{top1}{RGB}{198,239,206} 
\definecolor{top2}{RGB}{255,235,156} 

\title{MambaTrans: Multimodal Fusion Image Translation via Large Language Model Priors for Downstream Visual Tasks}

\author{
Yushen Xu$^{1}$  \quad \textbf{Xiaosong Li}$^{1,2}$\thanks{Corresponding author.} \quad \textbf{Zhenyu Kuang}$^2$ \quad \textbf{Xiaoqi Cheng}$^2$ \quad \textbf{Haishu Tan}$^{1}$ \quad \textbf{Huafeng Li}$^{3}$\\\\
$^1$School of Physics and Optoelectronic Engineering \\ Foshan University, Foshan, China\\
$^2$Guangdong-HongKong-Macao Joint Laboratory for Intelligent Micro-Nano Optoelectronic Technology\\ Foshan, China\\
$^3$School of Information Engineering and Automation \\ Kunming University of Science and Technology, Kunming, China\\
{\tt\small 2112355010@stu.fosu.edu.cn, lixiaosong@buaa.edu.cn, kmustkzy@126.com}\\
{\tt\small chengxiaoqi@fosu.edu.cn, tanghaishu@fosu.edu.cn, lhfchina99@kust.edu.cn } \\
}

\begin{document}

\maketitle

\begin{abstract}
The goal of multimodal image fusion is to integrate complementary information from infrared and visible images, generating multimodal fused images for downstream tasks. Existing downstream pre-training models are typically trained on visible images. However, the significant pixel distribution differences between visible and multimodal fusion images can degrade downstream task performance, sometimes even below that of using only visible images. This paper explores adapting multimodal fused images with significant modality differences to object detection and semantic segmentation models trained on visible images. To address this, we propose MambaTrans, a novel multimodal fusion image modality translator. MambaTrans uses descriptions from a multimodal large language model and masks from semantic segmentation models as input. Its core component, the Multi-Model State Space Block, combines mask-image-text cross-attention and a 3D-Selective Scan Module, enhancing pure visual capabilities. By leveraging object detection prior knowledge, MambaTrans minimizes detection loss during training and captures long-term dependencies among text, masks, and images. This enables favorable results in pre-trained models without adjusting their parameters. Experiments on public datasets show that MambaTrans effectively improves multimodal image performance in downstream tasks. 
\end{abstract}

\section{Introduction}
\begin{figure}[htbp]
\centering	
\includegraphics[width=1.0\linewidth]{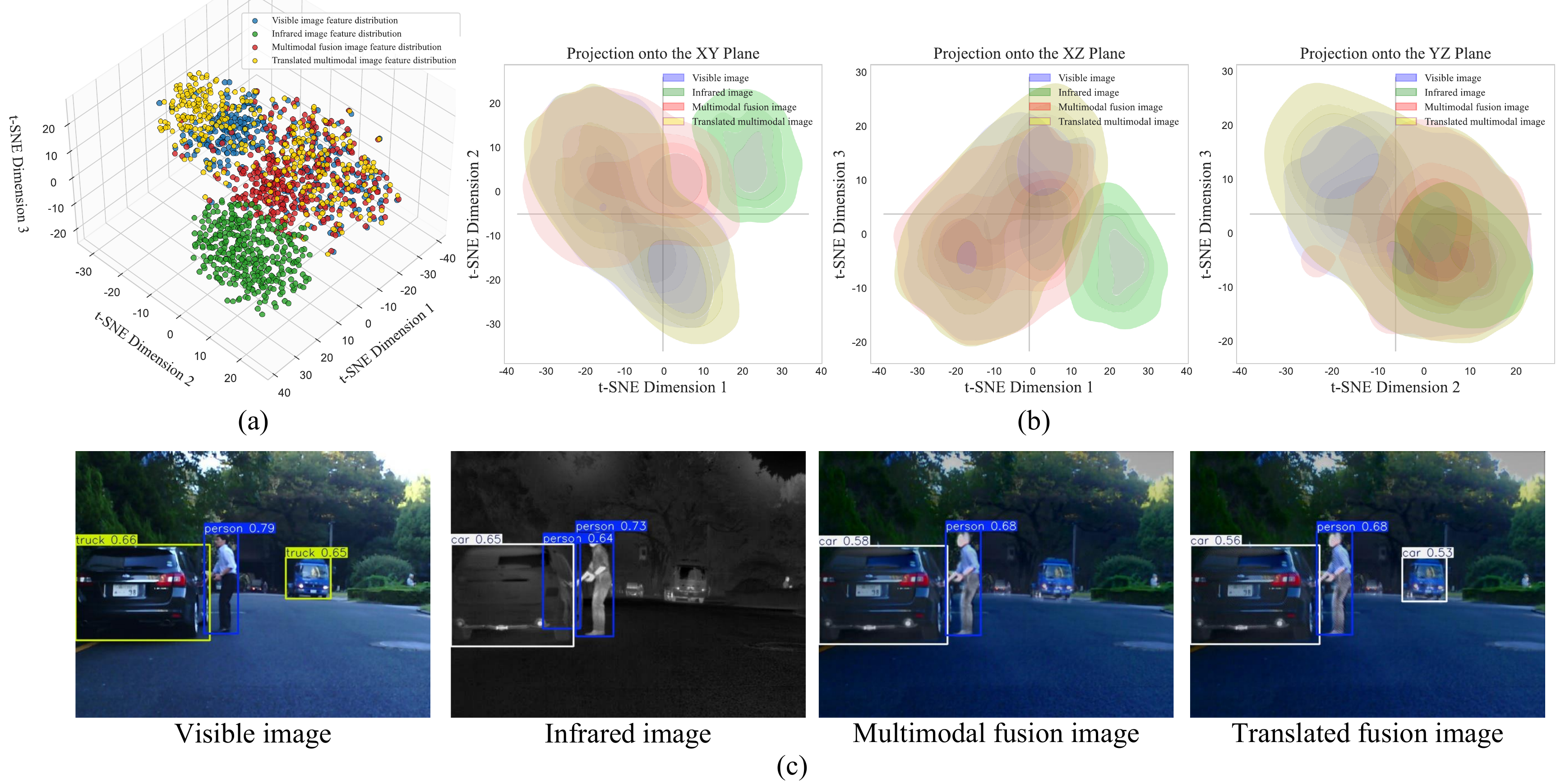}
\caption{(a) The 2048-dimensional features are extracted by using the pre-trained ResNet-50 and compressed to a three-dimensional space through the t-SNE algorithm \cite{6}. The data distribution of visible (blue), infrared (green), fused images (red), and translated images (yellow) is visualized. (b) It is a two-dimensional projection of the data distribution a. (c) It presents the performance discrepancies of different modality images in the detection task.}
\label{Fig1}
\end{figure}

Infrared-visible multimodal image fusion has significant application value in visual tasks such as night-time surveillance, autonomous driving, and security \cite{li2024deep,li2024all}. Infrared images can highlight the thermal information of targets under low-light conditions, while visible images contain rich texture and color details \cite{li2023infrared}. Effectively fusing the two is expected to significantly improve the performance of downstream visual tasks such as object detection and semantic segmentation \cite{2,xu2024simultaneous}. However, as shown in Figure \ref{Fig1}(a,b), there are significant differences in the distribution of fused images compared to visible images. Currently, pre-trained models widely used in object detection \cite{3} and semantic segmentation tasks \cite{4} are typically trained on large-scale natural image RGB datasets such as COCO \cite{5}. If one hopes to introduce fused images as input into downstream task models without weakening the knowledge of existing detectors, it is necessary to first bridge the distribution gap between domains caused by modality differences, making the fused images more similar to the image distribution learned by the pre-trained models.

In response to the above challenges, the image fusion and task collaborative optimization methods that have emerged in recent years \cite{7,8} attempt to guide the image fusion process through the prior knowledge of downstream tasks, thereby enhancing task performance. However, these methods typically fail to decouple image fusion from downstream task optimization, resulting in models that focus more on the performance of downstream tasks while neglecting the reconstruction quality of the fused images themselves. As shown in Figure \ref{Fig3}, this often leads to the visual quality of the fused images being inferior to that of traditional fusion methods.

To address the aforementioned issues, this paper proposes a multimodal image translation model named \textbf{MambaTrans}, which takes into account both image quality and the performance of downstream detection and segmentation tasks. MambaTrans takes the fused image as the main input and combines the text description and segmentation mask generated by a multimodal large language model \cite{jie2025fs} as auxiliary information to guide the modality translation process, generating an image representation that conforms to the visible light distribution. The model first extracts the initial features of the fused image and mask through convolutional layers and encodes the text using an encoder pre-trained on a large model. Subsequently, these features are input into multiple layers of multimodal state space blocks (MM-SSB) to achieve deep feature modeling and modality translation. MM-SSB is the core module of MambaTrans, supporting multimodal input and enabling semantic interaction among text, mask, and fused image through a multimodal attention interaction module. During the training phase, we incorporate the prior information of object detection into the loss function to minimize the object detection error. Through the above design, MambaTrans can bridge the modality gap, thereby activating the potential of the pre-trained model and significantly improving the accuracy of downstream tasks.

Our main contributions are as follows:
\begin{enumerate}
    \item This paper innovatively proposes MambaTrans, a modality translation framework for multimodal fusion images. This method adaptively adjusts the feature distribution of the fusion image to that of visible images, thereby better adapting to the pre-trained target detection and semantic segmentation models based on natural images, significantly improving the performance of downstream tasks.
    \item We designed the MM-SSB, which can simultaneously receive text descriptions, saliency masks, and fused images generated by multimodal large language models as input. By introducing the image-mask-text modality attention interaction module and 3D state space Module (3D-SSM), it effectively enhances the semantic representation ability of the fused images and enables the model to better learn the distribution characteristics of natural images.
    \item Experiments conducted on the two public multimodal datasets, MSRS and M\textsuperscript{3}FD, have demonstrated that the fused images processed by MambaTrans achieve performance improvements in both object detection and semantic segmentation tasks, while maintaining stable image visual quality with slight enhancements in relevant metrics. This verifies the proposed method's excellent balance between image quality and task performance. 
\end{enumerate}

\section{Related Work}
\subsection{Downstream Task-driven Multimodal Image Fusion}
Multimodal image fusion for downstream tasks is an important direction in the field of image fusion. Deep learning has promoted the introduction of task-guided mechanisms to achieve the collaborative optimization of fusion and performance. Tardal \cite{8} proposed a two-layer optimization formula to model the intrinsic relationship between object detection and image fusion and carried out the optimization process to construct a target-oriented two-layer learning network. TIMFusion\cite{9} integrates downstream task information into unsupervised learning of image fusion through constraint strategies and designs an efficient implicit search scheme to discover compact model architectures. SeAFusion \cite{10} combines image fusion with the segmentation task, introduces semantic loss and cyclic training to enhance the information content and semantic expression of the fusion results. MRFS\cite{11} is a coupled learning framework that achieves image fusion and segmentation through mutual reinforcement, improving visual quality and segmentation accuracy. The promoting effect of these methods on downstream tasks depends on their combined use with fusion algorithms and cannot decouple the fusion algorithm from the downstream task optimization algorithm, thereby limiting the further improvement of high-performance fusion algorithms in downstream tasks.

\subsection{Image Modality Translation}
The image modality translation task aims to map source domain images to the target domain while preserving their inherent content \cite{12,13}. Its core objective is to align the image distribution in the transformed domain with that of the target domain by finding a transformation function. InfraGAN proposed an image-level adaptation method focusing on RGB to IR conversion, which enhances the transformation effect by optimizing the image quality loss \cite{14}. Additionally, Herrmann et al. utilized traditional image preprocessing techniques to convert IR images to RGB images, thereby achieving the object detection task in the RGB/IR modalities without modifying parameters \cite{15}. HalluciDet \cite{16} improved the performance of object detection by introducing an image transformation mechanism, but it requires obtaining source RGB data from the same domain as the target domain in advance to pre-train the detector, which may impose certain limitations in practical applications. ModTr \cite{17} uses a small transformation network to adapt infrared input images, and this network is trained to directly minimize the detection loss. Current research mainly focuses on translating IR images to RGB images, and there is a lack of methods for translating multimodal fusion images to visible images. Moreover, existing methods struggle to optimize both segmentation and detection tasks simultaneously.

\section{Method}
This study aims to introduce the state space model \cite{18} (SSM) into the field of multimodal image translation. In this section, we first present the basic knowledge of SSM, then outline the multimodal translation model MambaTrans, and describe in detail the proposed MM-SSB, text-visual state space module (TV-SSM), and 3D selective scanning module (3D SSM). Finally, we explain how to incorporate the prior model of object detection into the loss function.

\subsection{Overall Architecture}
\begin{figure}[htbp]
\centering	
\includegraphics[width=1.0\linewidth]{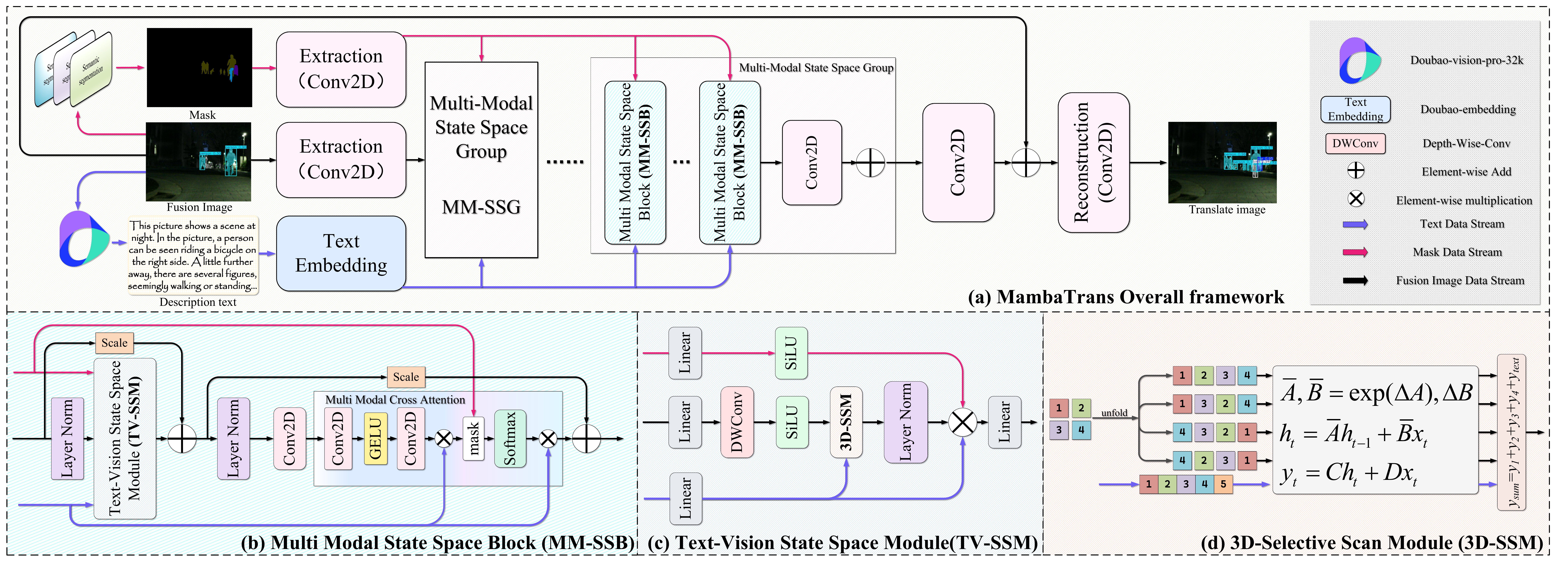}
\caption{The overall framework of our MambaTrans. MambaTrans translation processing significantly improves the performance of multimodal fusion images in detection and segmentation tasks.}
\label{Fig2}
\end{figure}

As shown in Figure \ref{Fig2}, MambaTrans converts multimodal fused images into images that conform to the visible light distribution. Its input includes infrared-visible light fusion images, segmentation masks, and image description texts. The image and mask are processed through convolutional layers to extract feature representations, while the text is encoded into semantic embeddings by a pre-trained large language model encoder. These features are then sent to a network composed of multiple multimodal state space groups (MM-SSG) for step-by-step fusion and transformation. Each MM-SSG contains several MM-SSB, which model multimodal interactions and spatiotemporal dependencies through a mask-image-text cross-modal attention mechanism and the TV-SSM module. Finally, the sequence features are remapped into two-dimensional image feature maps and the output image is reconstructed through convolution operations. During the training phase, the Task-Aware Charbonnier Loss (TAC Loss) is used to optimize the model, ensuring that the output image conforms to the characteristics of visible light distribution, while maintaining visual effects and incorporating target detection prior knowledge to enhance performance.

\subsection{Shallow Feature Extraction of Text-Semantic Images}
Before entering MM-SSB, MambaTrans performs shallow feature extraction on images, masks, and text separately. Images generate low-level visual feature maps with a channel dimension of 180 through convolutional layers; masks generate feature maps with the same dimension as image features through independent convolutional layers to ensure channel alignment between the two. Text generates high-level semantic feature sequences through a pre-trained large language model encoder. Subsequently, the image and mask feature maps are unfolded from two-dimensional H×W to one-dimensional patch sequences through patch embedding with a stride of 1, with each token corresponding to a feature vector at a spatial position. Ultimately, we obtain image feature sequences, mask feature sequences, and text feature sequences.

\subsection{Multimodal State Space Block}
Previously, the image processing network based on Mamba only supported single-modal image data processing. Given that Mamba originated from the field of natural language processing and was later transferred to the visual domain, we hypothesized that it could also perform well in simultaneously processing text and image data. To this end, we customized MM-SSB by adding text input to make up for the deficiency of the original visual SSM block in semantic feature expression.

Specifically, at the shallow layer, the image and mask features are processed into sequences of the identical length and can be regarded as aligned multimodal sequence inputs when entering MM-SSB. The text sequence, as supplementary information, is processed in parallel with the visual sequence and interactively fused. As illustrated in Figure \ref{Fig2}(b), the input fused image feature \(F_I\) is initially subjected to normalization and subsequently enters the Text-Vision State-Space Module (TV-SSM) along with the mask feature \(F_m\) and the text feature \(F_t\). The output is then added to the original fused image feature after normalization and adjustment by the scaling factor \(s \in \mathbb{R}^C\).
\begin{equation}
    Z^{n}=TV - SSM(LN(F_{I}),F_{mask},F_{text})+s\cdot LN(F_{I})
\end{equation}

Then, after undergoing normalization and passing through the convolutional layer, \(Z^{n}\), along with the mask feature \(F_m\) and the text feature \(F_t\), is fed into the modality attention interaction module. The multimodal cross attention module (MM-CA) is depicted in Figure \ref{Fig2}(b) and will be elaborated on specifically in the \ref{MMCA} section. The output of the MM-CA is added to the features of the previous stage to form the final output of this block:
\begin{equation}
    F^{n + 1} = MMCA(Conv(LN(Z^{n})), F_{mask}, F_{text}) + Z^{n}
\end{equation}

\subsection{Multimodal Cross Attention}
\label{MMCA}
The MM-CA introduces a multi-head attention mechanism, enabling direct interaction among text, image, and mask features and guiding the extraction and emphasis of features from one another. We have devised a ternary cross-attention strategy, concurrently considering the information flows of the mask, image, and text modalities: the text-image-mask fusion feature serves as the query, while the text feature sequence acts as the key and value for computing the attention weights. Specifically, the text-image-mask fusion feature undergoes \(1 \times 1\) and \(3 \times 3\) convolution transformations and is then fused in the channel dimension to construct the key/value sequence; the text description sequence is mapped to the same embedding dimension as the visual features and employed as the query sequence. The multi-head attention calculation generates the attention weights of the text query at each visual position, allowing the text semantics to focus on the corresponding image regions. Subsequently, the mask is utilized to mask the regions without the target to further enhance the target features.


\subsection{Text-Vision State-Space Module}
Inspired by the successful realization of long-range dependency modeling with linear complexity in visual state space modeling by Visual Mamba \cite{19,20,21}, we introduce the visual state space module into the multimodal image translation task and integrate text and mask inputs within this framework. As depicted in Figure \ref{Fig2}(c), the inputs comprise image features \(X \in \mathbb{R}^{H\times W\times C}\), mask features \(M \in \mathbb{R}^{H\times W\times C}\), and text features \(T \in \mathbb{R}^{H\times W\times C}\). Among them, the image features and mask features, as two parallel branches, are respectively transformed into a shared high-dimensional latent space via independent linear mapping layers. After the completion of feature projection, the image features undergo convolution operations and then enter the 3D Selective Scan module along with the text features. 

As depicted in Figure \ref{Fig2}(d), the two-dimensional image features are flattened into a one-dimensional sequence and scanned in four directions: from the top left corner to the bottom right corner, from the bottom right corner to the top left corner, from the top right corner to the bottom left corner, and from the bottom left corner to the top right corner. Meanwhile, the text features are processed through a one-dimensional unidirectional selective scan module, which can be regarded as conducting scanning in the third dimension. Hence, the overall process is termed 3D-Selective Scan: it encompasses bidirectional scanning within the two-dimensional image plane and one-dimensional text sequence scanning. Finally, the selectively scanned features from different spatial directions are element-wise added to the semantically rich text features and restored to their two-dimensional structure through reshaping operations, thereby accomplishing the integration of spatial information of the input data.
\begin{equation}
    y = y_1 + y_2 + y_3 + y_4 + y_{text}
\end{equation}
where, \(y_1,y_2,y_3,y_4\) represents the two-dimensional selective scanning output of the image from different spatial directions, while \(y_{text}\)represents the one-dimensional selective scanning output of the text.

The fused features subsequently undergo a nonlinear transformation via the semantic weights derived from the mask. The mask features, post activation by \(SiLU\) \cite{22}, undertake element-wise multiplication, thereby selectively highlighting the spatial regions associated with the semantic cues of the mask. Eventually, the output features are mapped back to the original image feature dimension through a normalization layer and a linear projection layer.

\subsection{Task-Aware Charbonnier Loss}
To ensure that the distribution of multimodal images approaches that of visible images as closely as possible and strikes a favorable balance between visual quality and the performance of downstream tasks, this research designs a task-aware loss function - TAC Loss. The construction of TAC Loss encompasses two core loss components: pixel-level reconstruction loss and task-oriented object detection loss. Firstly, the pixel-level reconstruction loss is defined using the Charbonnier loss function as follows:
\begin{equation}
    L_{\text{Charbonnier}}(\hat{I}, f, v)=\alpha\sum_{i = 1}^{n}\sqrt{(\hat{I}_i - f_i)^2 + \epsilon^2}+\beta\sum_{i = 1}^{n}\sqrt{(\hat{I}_i - v_i)^2 + \epsilon^2}
\end{equation}
where, $\hat{I}$ represents the image generated by the translator, $f$ represents the original multimodal image, $v$ represents the visible - light image, $n$ is the number of pixels in the image, $\epsilon$ is a small constant for stable calculation (usually taken as $10^{-3}$), and $\alpha, \beta$ are adjustment coefficients. 

This loss function can robustly handle minor pixel errors in image reconstruction and effectively avoid the problems of gradient explosion or vanishing. By adjusting the coefficient, it ensures that the translated image conforms to the visible light distribution while also taking into account the restoration of the visual effect of the fused image, and achieves a balance between the two.

Secondly, the task-oriented loss function of the pre-trained Faster R-CNN \cite{ren2015faster} is introduced to enhance the performance of the translated images in the object detection task. The loss function of Faster R-CNN consists of four components, which are defined as follows:
\begin{equation}
    L_{\text{detection}} = L_{\text{cls}} + L_{\text{bbox}} + L_{\text{obj}} + L_{\text{rpn}}
\end{equation}
where, $L_{\text{cls}}$ is the classification loss for object categories, $L_{\text{bbox}}$ is the bounding box regression loss, $L_{\text{obj}}$ is the objectness loss, and $L_{\text{rpn}}$ is the region proposal network (RPN) bounding box regression loss. 

Ultimately, we define the TAC Loss as the weighted summation of the two losses:
\begin{equation}
    L_{TAC} = \lambda L_{\text{Charbonnier}} + \theta L_{\text{detection}}
\end{equation}
where, \(\lambda\) and \(\theta\) are weight hyperparameters used to adjust the loss balance between pixel reconstruction and object detection tasks.

\section{Experiences}
\subsection{Experimental Settings}
\subsubsection{Dataset}
The experiments presented in this paper were carried out for training and testing on the M\textsuperscript{3}FD \cite{8} and MSRS \cite{23} datasets. Specifically, within the M\textsuperscript{3}FD dataset, 3,000 images were employed for training, 600 for validation, and another 600 for testing; whereas in the MSRS dataset, 1,083 images were utilized for training, 80 for validation, and 361 for testing. Both datasets furnish labels applicable to downstream tasks, among which MSRS contains semantic segmentation labels, and M\textsuperscript{3}FD contains object detection labels.
\subsubsection{Evaluation}
During the model training and validation process, to ensure the model's generalization ability, we utilized images generated by eight different fusion methods, including four task-driven methods (SAGE \cite{24}, DCEvo \cite{25}, TimFusion \cite{9} and CAF \cite{26}) and four data-driven methods (EMMA \cite{28}, CDDFuse \cite{29}, SHIP \cite{30} and CoCoNet \cite{31}). As the baseline comparison method for image translation, we selected CUT \cite{33}. In terms of performance evaluation, for object detection and semantic segmentation tasks, the mAP \cite{34} metric was adopted; while for the evaluation of image generation effects, this paper employed five metrics, including: entropy (EN) \cite{35}, average gradient (AG) \cite{36}, peak signal-to-noise ratio (PSNR), structural similarity (FSIM) \cite{37}, image quality evaluation index (\(Q_{TE}\)) \cite{38}, and spatial frequency (SF) \cite{39}. All indicators are such that the larger the value, the better the performance.
\subsubsection{Training Details}
We perform data augmentation through horizontal flipping and random cropping, while ensuring that the labels are consistent with the image transformations. During training, we use a batch size of 6 and conduct distributed training with 4 RTX 3090 GPUs. The optimizer is Adam with an initial learning rate of 1.000e-04, and the learning rate is halved at specific iteration milestones. The weight parameters in the loss function are set to 5 and 1 respectively. Since the official pre-trained model of CUT\cite{40} does not support the multimodal-to-visible light image translation task, we retrained the model and adjusted its batch size to be consistent with MambaTrans in the experiments to ensure fairness. The multimodal large language model used is Doubao-vision-pro-32k \cite{volcengine-doubao-vision-pro-32k}. Mask generation is achieved by using three state-of-the-art semantic segmentation models, FastInst \cite{he2023fastinst}, Cascade-Mask-RCNN-Swin \cite{cai2018cascade}, and MaskDINO-Swin \cite{li2023mask}, and the final mask is generated through a pixel-level majority voting strategy (at least two models' results are consistent).

We selected YOLOv11 \cite{khanam2024yolov11}, a widely used downstream task model, for detection and segmentation. Specifically, YOLOv11 was trained on the MSRS dataset's visible images for segmentation (8 categories) and on the M\textsuperscript{3}FD dataset's visible images for detection (6 categories). The training and test set division matched the earlier dataset section. For the tasks, the yolo11x version was used for detection, and the yolo11x-seg version for segmentation. Both were trained on 4 NVIDIA GeForce RTX 3090 GPUs with a batch size of 40 for 100 rounds.

\subsection{Comparison with Translation Approaches}
\subsubsection{Visual Effect Comparison}
\begin{figure}[h!]
\centering	
\includegraphics[width=1.0\linewidth]{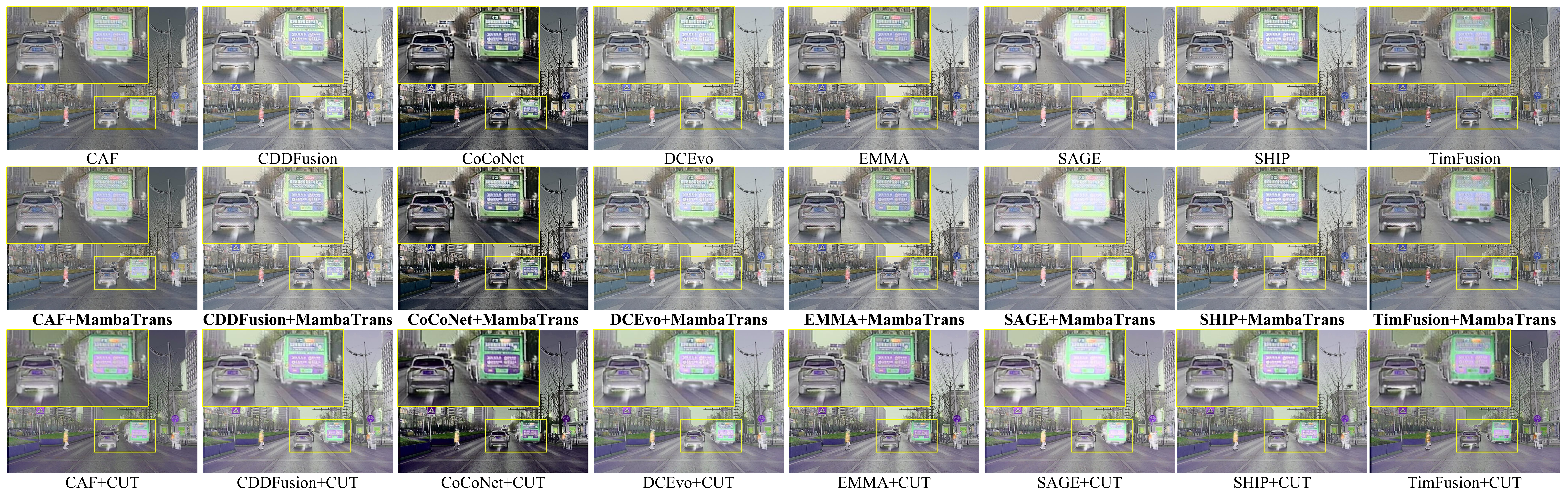}
\caption{Visual effect comparison.}
\label{Fig3}
\end{figure}
\begin{table}[h!]
\centering
\caption{The average objective metrics, CUT reconstruction results, and MambaTrans translation results of 8 fusion methods, \colorbox[HTML]{C6EFCE}{\color[HTML]{006100}green} indicating the best and \colorbox[HTML]{FFEB9C}{\color[HTML]{9C5700}yellow} indicating the second-best.}
\label{tab:1}
\resizebox{\textwidth}{!}{%
\begin{tabular}{c|ccccc|ccccc}
\hline
Dataset &
  \multicolumn{5}{c|}{MSRS} &
  \multicolumn{5}{c}{M\textsuperscript{3}FD} \\
Method &
  \(Q_{TE}\)$\uparrow$ &
  EN$\uparrow$ &
  SF$\uparrow$ &
  PSNR$\uparrow$ &
  FSIM$\uparrow$ &
  \(Q_{TE}\)$\uparrow$ &
  EN $\uparrow$&
  SF$\uparrow$ &
  PSNR$\uparrow$ &
  FSIM$\uparrow$ \\ \hline
\rowcolor[HTML]{F2F2F2} 
CAF (\textit{IJCAI'24}) &
  0.291 &
  6.041 &
  19.560 &
  17.434 &
  0.806 &
  0.337 &
  7.046 &
  14.401 &
  14.951 &
  0.819 \\
\rowcolor[HTML]{F2F2F2} 
CAF+CUT &
  0.288 &
  6.031 &
  9.820 &
  17.596 &
  0.797 &
  0.347 &
  6.962 &
  10.332 &
  14.777 &
  0.816 \\
\rowcolor[HTML]{C6EFCE} 
\cellcolor[HTML]{F2F2F2}\textbf{CAF+MambaTrans} &
  \cellcolor[HTML]{FFEB9C}{\color[HTML]{9C5700} 0.290} &
  {\color[HTML]{006100} 6.055} &
  \cellcolor[HTML]{FFEB9C}{\color[HTML]{9C5700} 18.720} &
  {\color[HTML]{006100} 17.492} &
  {\color[HTML]{006100} 0.807} &
  {\color[HTML]{006100} 0.356} &
  {\color[HTML]{006100} 7.144} &
  {\color[HTML]{006100} 14.778} &
  \cellcolor[HTML]{FFEB9C}{\color[HTML]{9C5700} 14.949} &
  {\color[HTML]{006100} 0.829} \\
CDDFusion (\textit{CVPR'23}) &
  0.429 &
  6.788 &
  11.535 &
  16.230 &
  0.828 &
  0.387 &
  7.071 &
  15.538 &
  12.882 &
  0.821 \\
CDDFusion+CUT &
  0.364 &
  6.779 &
  9.519 &
  16.638 &
  0.822 &
  0.364 &
  6.974 &
  11.734 &
  13.169 &
  0.817 \\
\textbf{CDDFusion+MambaTrans} &
  \cellcolor[HTML]{FFEB9C}{\color[HTML]{9C5700} 0.409} &
  \cellcolor[HTML]{C6EFCE}{\color[HTML]{006100} 6.788} &
  \cellcolor[HTML]{C6EFCE}{\color[HTML]{006100} 12.035} &
  \cellcolor[HTML]{C6EFCE}{\color[HTML]{006100} 16.240} &
  \cellcolor[HTML]{C6EFCE}{\color[HTML]{006100} 0.829} &
  \cellcolor[HTML]{FFEB9C}{\color[HTML]{9C5700} 0.380} &
  \cellcolor[HTML]{C6EFCE}{\color[HTML]{006100} 7.079} &
  \cellcolor[HTML]{C6EFCE}{\color[HTML]{006100} 15.994} &
  \cellcolor[HTML]{C6EFCE}{\color[HTML]{006100} 13.882} &
  \cellcolor[HTML]{C6EFCE}{\color[HTML]{006100} 0.821} \\
\rowcolor[HTML]{F2F2F2} 
CoCoNet (\textit{IJCV'24})&
  0.530 &
  7.667 &
  22.165 &
  9.710 &
  0.632 &
  0.333 &
  7.768 &
  25.164 &
  12.155 &
  0.723 \\
\rowcolor[HTML]{F2F2F2} 
CoCoNet+CUT &
  0.526 &
  7.649 &
  16.149 &
  10.038 &
  0.639 &
  0.330 &
  7.714 &
  18.683 &
  12.143 &
  0.724 \\
\rowcolor[HTML]{C6EFCE} 
\cellcolor[HTML]{F2F2F2}\textbf{CoCoNet+MambaTrans} &
  {\color[HTML]{006100} 0.531} &
  {\color[HTML]{006100} 7.676} &
  {\color[HTML]{006100} 22.699} &
  {\color[HTML]{006100} 9.722} &
  \cellcolor[HTML]{FFEB9C}{\color[HTML]{9C5700} 0.636} &
  {\color[HTML]{006100} 0.341} &
  {\color[HTML]{006100} 7.776} &
  {\color[HTML]{006100} 25.334} &
  {\color[HTML]{006100} 12.172} &
  {\color[HTML]{006100} 0.725} \\
DCEvo (\textit{CVPR'25}) &
  0.385 &
  6.720 &
  11.459 &
  16.394 &
  0.831 &
  0.394 &
  7.033 &
  15.009 &
  12.668 &
  0.824 \\
DCEvo+CUT &
  0.340 &
  6.711 &
  9.444 &
  16.793 &
  0.824 &
  0.366 &
  6.900 &
  11.465 &
  13.058 &
  0.821 \\
\textbf{DCEvo+MambaTrans} &
  \cellcolor[HTML]{FFEB9C}{\color[HTML]{9C5700} 0.374} &
  \cellcolor[HTML]{C6EFCE}{\color[HTML]{006100} 6.723} &
  \cellcolor[HTML]{C6EFCE}{\color[HTML]{006100} 11.916} &
  \cellcolor[HTML]{C6EFCE}{\color[HTML]{006100} 16.410} &
  \cellcolor[HTML]{C6EFCE}{\color[HTML]{006100} 0.832} &
  \cellcolor[HTML]{FFEB9C}{\color[HTML]{9C5700} 0.386} &
  \cellcolor[HTML]{C6EFCE}{\color[HTML]{006100} 7.031} &
  \cellcolor[HTML]{C6EFCE}{\color[HTML]{006100} 15.508} &
  \cellcolor[HTML]{FFEB9C}{\color[HTML]{9C5700} 12.670} &
  \cellcolor[HTML]{C6EFCE}{\color[HTML]{006100} 0.824} \\
\rowcolor[HTML]{F2F2F2} 
EMMA (\textit{CVPR'24}) &
  0.382 &
  6.801 &
  11.624 &
  16.248 &
  0.822 &
  0.366 &
  7.128 &
  16.013 &
  12.961 &
  0.818 \\
\rowcolor[HTML]{F2F2F2} 
EMMA+CUT &
  0.347 &
  6.784 &
  9.838 &
  16.609 &
  0.817 &
  0.348 &
  7.007 &
  12.745 &
  13.212 &
  0.816 \\
\rowcolor[HTML]{C6EFCE} 
\cellcolor[HTML]{F2F2F2}\textbf{EMMA+MambaTrans} &
  \cellcolor[HTML]{FFEB9C}{\color[HTML]{9C5700} 0.373} &
  {\color[HTML]{006100} 6.801} &
  {\color[HTML]{006100} 12.135} &
  {\color[HTML]{006100} 16.257} &
  {\color[HTML]{006100} 0.824} &
  {\color[HTML]{006100} 0.368} &
  {\color[HTML]{006100} 7.225} &
  {\color[HTML]{006100} 16.634} &
  \cellcolor[HTML]{FFEB9C}{\color[HTML]{9C5700} 13.959} &
  {\color[HTML]{006100} 0.819} \\
SAGE (\textit{CVPR'24})&
  0.424 &
  6.285 &
  10.410 &
  17.154 &
  0.824 &
  0.382 &
  7.014 &
  13.489 &
  13.029 &
  0.818 \\
SAGE+CUT &
  0.402 &
  6.329 &
  8.543 &
  17.494 &
  0.814 &
  0.366 &
  6.889 &
  10.533 &
  13.273 &
  0.814 \\
\textbf{SAGE+MambaTrans} &
  \cellcolor[HTML]{C6EFCE}{\color[HTML]{006100} 0.491} &
  \cellcolor[HTML]{C6EFCE}{\color[HTML]{006100} 6.289} &
  \cellcolor[HTML]{C6EFCE}{\color[HTML]{006100} 10.872} &
  \cellcolor[HTML]{C6EFCE}{\color[HTML]{006100} 17.155} &
  \cellcolor[HTML]{C6EFCE}{\color[HTML]{006100} 0.825} &
  \cellcolor[HTML]{C6EFCE}{\color[HTML]{006100} 0.385} &
  \cellcolor[HTML]{FFEB9C}{\color[HTML]{9C5700} 7.012} &
  \cellcolor[HTML]{C6EFCE}{\color[HTML]{006100} 14.076} &
  \cellcolor[HTML]{C6EFCE}{\color[HTML]{006100} 13.127} &
  \cellcolor[HTML]{C6EFCE}{\color[HTML]{006100} 0.818} \\
\rowcolor[HTML]{F2F2F2} 
SHIP (\textit{CVPR'24})&
  0.478 &
  6.618 &
  11.803 &
  15.933 &
  0.823 &
  0.390 &
  7.025 &
  16.162 &
  12.596 &
  0.824 \\
\rowcolor[HTML]{F2F2F2} 
SHIP+CUT &
  0.410 &
  6.631 &
  9.520 &
  16.385 &
  0.817 &
  0.361 &
  6.875 &
  11.948 &
  12.498 &
  0.821 \\
\rowcolor[HTML]{C6EFCE} 
\cellcolor[HTML]{F2F2F2}\textbf{SHIP+MambaTrans} &
  {\color[HTML]{006100} 0.477} &
  {\color[HTML]{006100} 6.618} &
  {\color[HTML]{006100} 12.251} &
  {\color[HTML]{006100} 15.944} &
  {\color[HTML]{006100} 0.825} &
  {\color[HTML]{006100} 0.397} &
  {\color[HTML]{006100} 7.121} &
  {\color[HTML]{006100} 16.636} &
  {\color[HTML]{006100} 12.696} &
  {\color[HTML]{006100} 0.825} \\
TimFusion (\textit{TPAMI'24})&
  0.549 &
  7.185 &
  11.494 &
  11.408 &
  0.723 &
  0.339 &
  7.043 &
  12.917 &
  14.920 &
  0.813 \\
TimFusion+CUT &
  0.522 &
  7.187 &
  9.245 &
  11.736 &
  0.723 &
  0.337 &
  6.991 &
  10.505 &
  14.764 &
  0.811 \\
\textbf{TimFusion+MambaTrans} &
  \cellcolor[HTML]{C6EFCE}{\color[HTML]{006100} 0.556} &
  \cellcolor[HTML]{C6EFCE}{\color[HTML]{006100} 7.188} &
  \cellcolor[HTML]{C6EFCE}{\color[HTML]{006100} 11.941} &
  \cellcolor[HTML]{FFEB9C}{\color[HTML]{9C5700} 11.413} &
  \cellcolor[HTML]{C6EFCE}{\color[HTML]{006100} 0.726} &
  \cellcolor[HTML]{C6EFCE}{\color[HTML]{006100} 0.341} &
  \cellcolor[HTML]{C6EFCE}{\color[HTML]{006100} 7.045} &
  \cellcolor[HTML]{C6EFCE}{\color[HTML]{006100} 13.639} &
  \cellcolor[HTML]{FFEB9C}{\color[HTML]{9C5700} 14.913} &
  \cellcolor[HTML]{C6EFCE}{\color[HTML]{006100} 0.814} \\ \hline
\end{tabular}%
}
\end{table}
Figure \ref{Fig3} presents the experimental results of the proposed method on the task of infrared and visible light image fusion. From left to right are: the fusion results of eight existing methods, the reconstruction results of the CUT method, and the translation results of MambaTrans. It can be seen that the existing methods have insufficient clarity at the boundaries of the main target areas and fail to enhance important targets such as pedestrians and cyclists specifically, resulting in insufficient positioning information. The images reconstructed by the CUT model have color shift issues, which affect the performance of downstream tasks. In contrast, the translation results of MambaTrans significantly enhance the features of pedestrians and vehicles, making their textures clearer and boundaries more distinct, which is beneficial for detection and segmentation models to accurately locate the targets.

The objective indicators are shown in Table \ref{tab:1}. Clearly, the fusion results obtained by using the translation model proposed in this paper are superior to the original methods in almost all fusion metrics. This fully validates the effectiveness of the proposed method in enhancing the fusion performance.

\begin{table}[h!]
\centering
\caption{The mAP values of semantic segmentation task on MSRS and object detection task on M\textsuperscript{3}FD, with \colorbox[HTML]{C6EFCE}{\color[HTML]{006100}green} indicating the best.}
\label{tab:23}
\resizebox{\textwidth}{!}{%
\begin{tabular}{c|cccccccccc|cccccccc}
\hline
Dataset &
  \multicolumn{10}{c|}{MSRS} &
  \multicolumn{8}{c}{M\textsuperscript{3}FD} \\
\textbf{Method} &
  \textbf{Car} &
  \textbf{Person} &
  \textbf{Bike} &
  \textbf{Curve} &
  \textbf{Car Stop} &
  \textbf{Guardrail} &
  \textbf{Color Cone} &
  \textbf{Bump} &
  \textbf{mAP50} &
  \textbf{mAP50-95} &
  \textbf{Person} &
  \textbf{Car} &
  \textbf{Bus} &
  \textbf{Lamp} &
  \textbf{Motorcycle} &
  \textbf{Truck} &
  \textbf{mAP50} &
  \textbf{mAP50-95} \\ \hline
VI &
  0.905 &
  0.587 &
  0.549 &
  0.737 &
  0.798 &
  0.973 &
  0.729 &
  0.763 &
  0.7551 &
  0.5 &
  0.969 &
  0.913 &
  0.923 &
  0.892 &
  0.699 &
  0.903 &
  0.8832 &
  0.581 \\
IR &
  0.221 &
  0.462 &
  0.147 &
  0.0437 &
  0.118 &
  0.0538 &
  0.148 &
  0.106 &
  0.1624 &
  0.162 &
  0.815 &
  0.795 &
  0.866 &
  0.283 &
  0.535 &
  0.707 &
  0.6660 &
  0.416 \\
\rowcolor[HTML]{F2F2F2} CAF (\textit{IJCAI'24}) &
  0.905 &
  0.705 &
  0.586 &
  0.74 &
  0.694 &
  0.995 &
  0.719 &
  0.817 &
  0.7701 &
  0.446 &
  0.899 &
  0.933 &
  0.943 &
  0.874 &
  0.762 &
  0.903 &
  0.8857 &
  0.632 \\
\rowcolor[HTML]{F2F2F2} CAF+CUT &
  0.851 &
  0.647 &
  0.501 &
  0.575 &
  0.56 &
  0.848 &
  0.565 &
  0.553 &
  0.6375 &
  0.352 &
  0.887 &
  0.907 &
  0.914 &
  0.77 &
  0.658 &
  0.866 &
  0.8337 &
  0.59 \\
\rowcolor[HTML]{C6EFCE} \cellcolor[HTML]{F2F2F2}\textbf{CAF+MambaTrans} &
  \color[HTML]{006100} 0.908 &
  \color[HTML]{006100} 0.723 &
  \color[HTML]{006100} 0.598 &
  \color[HTML]{006100} 0.745 &
  \color[HTML]{006100} 0.697 &
  \color[HTML]{006100} 0.995 &
  \color[HTML]{006100} 0.725 &
  \color[HTML]{006100} 0.826 &
  \color[HTML]{006100} 0.7771 &
  \color[HTML]{006100} 0.449 &
  \color[HTML]{006100} 0.904 &
  \color[HTML]{006100} 0.943 &
  \color[HTML]{006100} 0.949 &
  \color[HTML]{006100} 0.881 &
  \color[HTML]{006100} 0.793 &
  \color[HTML]{006100} 0.908 &
  \color[HTML]{006100} 0.8963 &
  \color[HTML]{006100} 0.643 \\
CDDFusion (\textit{CVPR'23}) &
  0.903 &
  0.723 &
  0.599 &
  0.754 &
  0.686 &
  0.988 &
  0.709 &
  0.893 &
  0.7819 &
  0.467 &
  0.894 &
  0.936 &
  0.941 &
  0.883 &
  0.753 &
  0.91 &
  0.8862 &
  0.634 \\
CDDFusion+CUT &
  0.882 &
  0.674 &
  0.519 &
  0.628 &
  0.6 &
  0.92 &
  0.62 &
  0.638 &
  0.6851 &
  0.397 &
  0.887 &
  0.907 &
  0.911 &
  0.815 &
  0.68 &
  0.863 &
  0.8438 &
  0.598 \\
\textbf{CDDFusion+MambaTrans} &
  \cellcolor[HTML]{C6EFCE}{\color[HTML]{006100} 0.907} &
  \cellcolor[HTML]{C6EFCE}{\color[HTML]{006100} 0.74} &
  \cellcolor[HTML]{C6EFCE}{\color[HTML]{006100} 0.614} &
  \cellcolor[HTML]{C6EFCE}{\color[HTML]{006100} 0.759} &
  \cellcolor[HTML]{C6EFCE}{\color[HTML]{006100} 0.757} &
  \cellcolor[HTML]{C6EFCE}{\color[HTML]{006100} 0.988} &
  \cellcolor[HTML]{C6EFCE}{\color[HTML]{006100} 0.716} &
  \cellcolor[HTML]{C6EFCE}{\color[HTML]{006100} 0.877} &
  \cellcolor[HTML]{C6EFCE}{\color[HTML]{006100} 0.7948} &
  \cellcolor[HTML]{C6EFCE}{\color[HTML]{006100} 0.464} &
  \cellcolor[HTML]{C6EFCE}{\color[HTML]{006100} 0.895} &
  \cellcolor[HTML]{C6EFCE}{\color[HTML]{006100} 0.944} &
  \cellcolor[HTML]{C6EFCE}{\color[HTML]{006100} 0.963} &
  \cellcolor[HTML]{C6EFCE}{\color[HTML]{006100} 0.884} &
  \cellcolor[HTML]{C6EFCE}{\color[HTML]{006100} 0.786} &
  \cellcolor[HTML]{C6EFCE}{\color[HTML]{006100} 0.921} &
  \cellcolor[HTML]{C6EFCE}{\color[HTML]{006100} 0.8988} &
  \cellcolor[HTML]{C6EFCE}{\color[HTML]{006100} 0.648} \\
\rowcolor[HTML]{F2F2F2} CoCoNet (IJCV'24) &
  0.896 &
  0.682 &
  0.565 &
  0.754 &
  0.673 &
  0.995 &
  0.701 &
  0.789 &
  0.7569 &
  0.453 &
  0.886 &
  0.931 &
  0.929 &
  0.875 &
  0.733 &
  0.905 &
  0.8765 &
  0.626 \\
\rowcolor[HTML]{F2F2F2} CoCoNet+CUT &
  0.866 &
  0.637 &
  0.529 &
  0.588 &
  0.584 &
  0.848 &
  0.609 &
  0.558 &
  0.6524 &
  0.366 &
  0.863 &
  0.906 &
  0.878 &
  0.802 &
  0.641 &
  0.84 &
  0.8217 &
  0.58 \\
\rowcolor[HTML]{C6EFCE} \cellcolor[HTML]{F2F2F2}\textbf{CoCoNet+MambaTrans} &
  \color[HTML]{006100} 0.899 &
  \color[HTML]{006100} 0.707 &
  \color[HTML]{006100} 0.595 &
  \color[HTML]{006100} 0.758 &
  \color[HTML]{006100} 0.666 &
  \color[HTML]{006100} 0.995 &
  \color[HTML]{006100} 0.688 &
  \color[HTML]{006100} 0.796 &
  \color[HTML]{006100} 0.7630 &
  \color[HTML]{006100} 0.457 &
  \color[HTML]{006100} 0.887 &
  \color[HTML]{006100} 0.941 &
  \color[HTML]{006100} 0.951 &
  \color[HTML]{006100} 0.885 &
  \color[HTML]{006100} 0.79 &
  \color[HTML]{006100} 0.915 &
  \color[HTML]{006100} 0.8948 &
  \color[HTML]{006100} 0.64 \\
DCEvo (\textit{CVPR'25})&
  0.908 &
  0.723 &
  0.591 &
  0.764 &
  0.678 &
  0.988 &
  0.723 &
  0.85 &
  0.7781 &
  0.466 &
  0.896 &
  0.936 &
  0.94 &
  0.883 &
  0.745 &
  0.914 &
  0.8857 &
  0.635 \\
DCEvo+CUT &
  0.877 &
  0.67 &
  0.527 &
  0.604 &
  0.614 &
  0.906 &
  0.62 &
  0.642 &
  0.6825 &
  0.394 &
  0.888 &
  0.909 &
  0.895 &
  0.812 &
  0.673 &
  0.857 &
  0.8390 &
  0.598 \\
\textbf{DCEvo+MambaTrans} &
  \cellcolor[HTML]{C6EFCE}{\color[HTML]{006100} 0.916} &
  \cellcolor[HTML]{C6EFCE}{\color[HTML]{006100} 0.741} &
  \cellcolor[HTML]{C6EFCE}{\color[HTML]{006100} 0.595} &
  \cellcolor[HTML]{C6EFCE}{\color[HTML]{006100} 0.778} &
  \cellcolor[HTML]{C6EFCE}{\color[HTML]{006100} 0.69} &
  \cellcolor[HTML]{C6EFCE}{\color[HTML]{006100} 0.989} &
  \cellcolor[HTML]{C6EFCE}{\color[HTML]{006100} 0.731} &
  \cellcolor[HTML]{C6EFCE}{\color[HTML]{006100} 0.855} &
  \cellcolor[HTML]{C6EFCE}{\color[HTML]{006100} 0.7869} &
  \cellcolor[HTML]{C6EFCE}{\color[HTML]{006100} 0.469} &
  \cellcolor[HTML]{C6EFCE}{\color[HTML]{006100} 0.904} &
  \cellcolor[HTML]{C6EFCE}{\color[HTML]{006100} 0.943} &
  \cellcolor[HTML]{C6EFCE}{\color[HTML]{006100} 0.953} &
  \cellcolor[HTML]{C6EFCE}{\color[HTML]{006100} 0.887} &
  \cellcolor[HTML]{C6EFCE}{\color[HTML]{006100} 0.786} &
  \cellcolor[HTML]{C6EFCE}{\color[HTML]{006100} 0.924} &
  \cellcolor[HTML]{C6EFCE}{\color[HTML]{006100} 0.8995} &
  \cellcolor[HTML]{C6EFCE}{\color[HTML]{006100} 0.646} \\
\rowcolor[HTML]{F2F2F2} EMMA (\textit{CVPR'24}) &
  0.912 &
  0.713 &
  0.593 &
  0.76 &
  0.679 &
  0.988 &
  0.732 &
  0.879 &
  0.7820 &
  0.465 &
  0.891 &
  0.934 &
  0.939 &
  0.876 &
  0.758 &
  0.906 &
  0.8840 &
  0.632 \\
\rowcolor[HTML]{F2F2F2} EMMA+CUT &
  0.883 &
  0.67 &
  0.512 &
  0.61 &
  0.593 &
  0.887 &
  0.609 &
  0.562 &
  0.6658 &
  0.387 &
  0.884 &
  0.905 &
  0.906 &
  0.823 &
  0.675 &
  0.864 &
  0.8428 &
  0.6 \\
\rowcolor[HTML]{C6EFCE} \cellcolor[HTML]{F2F2F2}\textbf{EMMA+MambaTrans} &
  \color[HTML]{006100} 0.918 &
  \color[HTML]{006100} 0.728 &
  \color[HTML]{006100} 0.6 &
  \color[HTML]{006100} 0.767 &
  \color[HTML]{006100} 0.697 &
  \color[HTML]{006100} 0.995 &
  \color[HTML]{006100} 0.744 &
  \color[HTML]{006100} 0.889 &
  \color[HTML]{006100} 0.7923 &
  \color[HTML]{006100} 0.465 &
  \color[HTML]{006100} 0.895 &
  \color[HTML]{006100} 0.942 &
  \color[HTML]{006100} 0.953 &
  \color[HTML]{006100} 0.883 &
  \color[HTML]{006100} 0.786 &
  \color[HTML]{006100} 0.924 &
  \color[HTML]{006100} 0.8972 &
  \color[HTML]{006100} 0.647 \\
SAGE (\textit{CVPR'24})&
  0.904 &
  0.699 &
  0.586 &
  0.747 &
  0.673 &
  0.98 &
  0.708 &
  0.84 &
  0.7671 &
  0.458 &
  0.899 &
  0.934 &
  0.94 &
  0.872 &
  0.752 &
  0.91 &
  0.8845 &
  0.636 \\
SAGE+CUT &
  0.857 &
  0.645 &
  0.497 &
  0.581 &
  0.588 &
  0.945 &
  0.589 &
  0.579 &
  0.6601 &
  0.378 &
  0.894 &
  0.913 &
  0.927 &
  0.818 &
  0.645 &
  0.891 &
  0.8480 &
  0.607 \\
\textbf{SAGE+MambaTrans} &
  \cellcolor[HTML]{C6EFCE}{\color[HTML]{006100} 0.911} &
  \cellcolor[HTML]{C6EFCE}{\color[HTML]{006100} 0.722} &
  \cellcolor[HTML]{C6EFCE}{\color[HTML]{006100} 0.598} &
  \cellcolor[HTML]{C6EFCE}{\color[HTML]{006100} 0.748} &
  \cellcolor[HTML]{C6EFCE}{\color[HTML]{006100} 0.688} &
  \cellcolor[HTML]{C6EFCE}{\color[HTML]{006100} 0.988} &
  \cellcolor[HTML]{C6EFCE}{\color[HTML]{006100} 0.719} &
  \cellcolor[HTML]{C6EFCE}{\color[HTML]{006100} 0.849} &
  \cellcolor[HTML]{C6EFCE}{\color[HTML]{006100} 0.7779} &
  \cellcolor[HTML]{C6EFCE}{\color[HTML]{006100} 0.462} &
  \cellcolor[HTML]{C6EFCE}{\color[HTML]{006100} 0.901} &
  \cellcolor[HTML]{C6EFCE}{\color[HTML]{006100} 0.94} &
  \cellcolor[HTML]{C6EFCE}{\color[HTML]{006100} 0.955} &
  \cellcolor[HTML]{C6EFCE}{\color[HTML]{006100} 0.879} &
  \cellcolor[HTML]{C6EFCE}{\color[HTML]{006100} 0.789} &
  \cellcolor[HTML]{C6EFCE}{\color[HTML]{006100} 0.926} &
  \cellcolor[HTML]{C6EFCE}{\color[HTML]{006100} 0.8983} &
  \cellcolor[HTML]{C6EFCE}{\color[HTML]{006100} 0.648} \\
\rowcolor[HTML]{F2F2F2} SHIP (\textit{CVPR'24})&
  0.904 &
  0.72 &
  0.602 &
  0.753 &
  0.683 &
  0.98 &
  0.718 &
  0.833 &
  0.7741 &
  0.463 &
  0.887 &
  0.929 &
  0.934 &
  0.874 &
  0.743 &
  0.907 &
  0.8790 &
  0.63 \\
\rowcolor[HTML]{F2F2F2} SHIP+CUT &
  0.872 &
  0.678 &
  0.5 &
  0.608 &
  0.607 &
  0.928 &
  0.611 &
  0.637 &
  0.6801 &
  0.393 &
  0.886 &
  0.905 &
  0.889 &
  0.808 &
  0.678 &
  0.843 &
  0.8348 &
  0.591 \\
\rowcolor[HTML]{C6EFCE} \cellcolor[HTML]{F2F2F2}\textbf{SHIP+MambaTrans} &
  \color[HTML]{006100} 0.907 &
  \color[HTML]{006100} 0.741 &
  \color[HTML]{006100} 0.59 &
  \color[HTML]{006100} 0.748 &
  \color[HTML]{006100} 0.694 &
  \color[HTML]{006100} 0.972 &
  \color[HTML]{006100} 0.693 &
  \color[HTML]{006100} 0.832 &
  \color[HTML]{006100} 0.7721 &
  \color[HTML]{006100} 0.465 &
  \color[HTML]{006100} 0.898 &
  \color[HTML]{006100} 0.943 &
  \color[HTML]{006100} 0.951 &
  \color[HTML]{006100} 0.882 &
  \color[HTML]{006100} 0.785 &
  \color[HTML]{006100} 0.918 &
  \color[HTML]{006100} 0.8962 &
  \color[HTML]{006100} 0.642 \\
TimFusion (\textit{TPAMI'24})&
  0.894 &
  0.679 &
  0.552 &
  0.728 &
  0.674 &
  0.995 &
  0.667 &
  0.866 &
  0.7569 &
  0.446 &
  0.878 &
  0.931 &
  0.941 &
  0.883 &
  0.75 &
  0.904 &
  0.8812 &
  0.624 \\
TimFusion+CUT &
  0.864 &
  0.635 &
  0.467 &
  0.555 &
  0.564 &
  0.963 &
  0.542 &
  0.589 &
  0.6474 &
  0.365 &
  0.864 &
  0.901 &
  0.904 &
  0.796 &
  0.638 &
  0.846 &
  0.8248 &
  0.577 \\
\textbf{TimFusion+MambaTrans} &
  \cellcolor[HTML]{C6EFCE}{\color[HTML]{006100} 0.896} &
  \cellcolor[HTML]{C6EFCE}{\color[HTML]{006100} 0.713} &
  \cellcolor[HTML]{C6EFCE}{\color[HTML]{006100} 0.562} &
  \cellcolor[HTML]{C6EFCE}{\color[HTML]{006100} 0.725} &
  \cellcolor[HTML]{C6EFCE}{\color[HTML]{006100} 0.651} &
  \cellcolor[HTML]{C6EFCE}{\color[HTML]{006100} 0.988} &
  \cellcolor[HTML]{C6EFCE}{\color[HTML]{006100} 0.68} &
  \cellcolor[HTML]{C6EFCE}{\color[HTML]{006100} 0.857} &
  \cellcolor[HTML]{C6EFCE}{\color[HTML]{006100} 0.7590} &
  \cellcolor[HTML]{C6EFCE}{\color[HTML]{006100} 0.449} &
  \cellcolor[HTML]{C6EFCE}{\color[HTML]{006100} 0.887} &
  \cellcolor[HTML]{C6EFCE}{\color[HTML]{006100} 0.941} &
  \cellcolor[HTML]{C6EFCE}{\color[HTML]{006100} 0.957} &
  \cellcolor[HTML]{C6EFCE}{\color[HTML]{006100} 0.885} &
  \cellcolor[HTML]{C6EFCE}{\color[HTML]{006100} 0.771} &
  \cellcolor[HTML]{C6EFCE}{\color[HTML]{006100} 0.91} &
  \cellcolor[HTML]{C6EFCE}{\color[HTML]{006100} 0.8918} &
  \cellcolor[HTML]{C6EFCE}{\color[HTML]{006100} 0.64} \\ \hline
\end{tabular}%
}
\end{table}
\subsubsection{Semantic Segmentation Effect Comparison}
\begin{figure}[htbp]
\centering	
\includegraphics[width=1.0\linewidth]{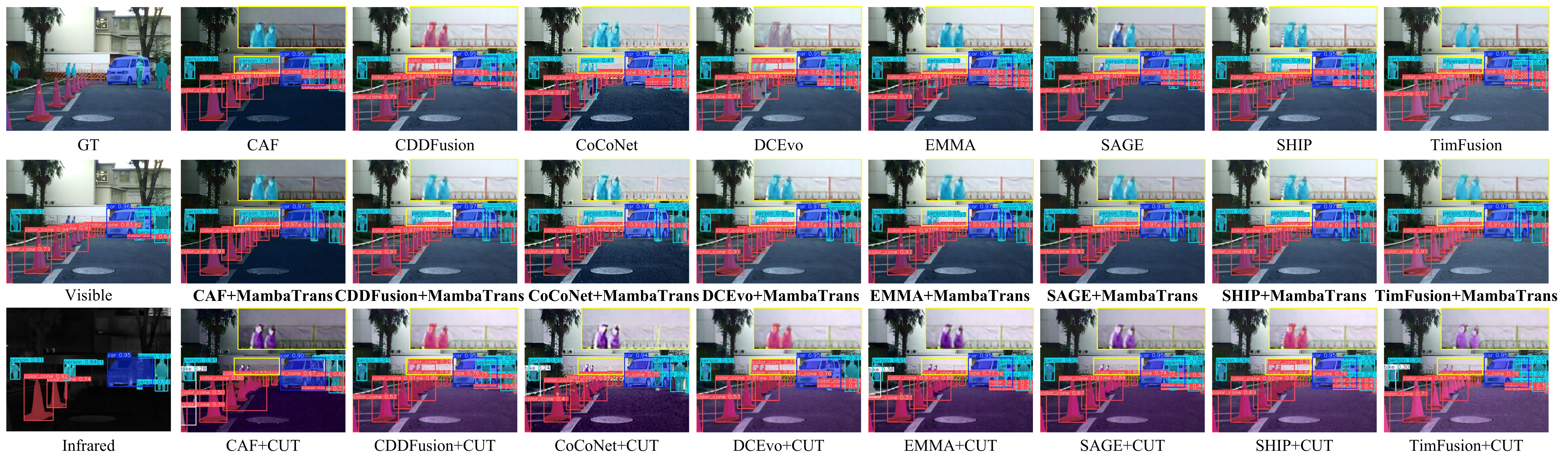}
\caption{Comparison of Segmentation Results of YOLOv11.}
\label{Fig4}
\end{figure}

This section conducts semantic segmentation experiments. Figure \ref{Fig4} presents two sets of subjective results. In the images generated by the existing fusion methods and the CUT reconstructed images, the person behind the traffic cone cannot be effectively segmented and recognized by YOLO v11; however, the image translated by MambaTrans enables YOLO v11 to accurately complete the task. Table \ref{tab:23} shows the specific segmentation metrics. The segmentation performance of visible light images is superior to that of most fusion images, indicating that the significant difference in modality distribution significantly affects the performance of the segmentation model. The MambaTrans proposed in this paper can effectively improve the segmentation performance of fusion images, achieving improvements in all categories, and some translation results even outperform the segmentation metrics of visible light images.

\subsubsection{Object Detection Effect Comparison}

\begin{figure}[h!]
\centering	
\includegraphics[width=1.0\linewidth]{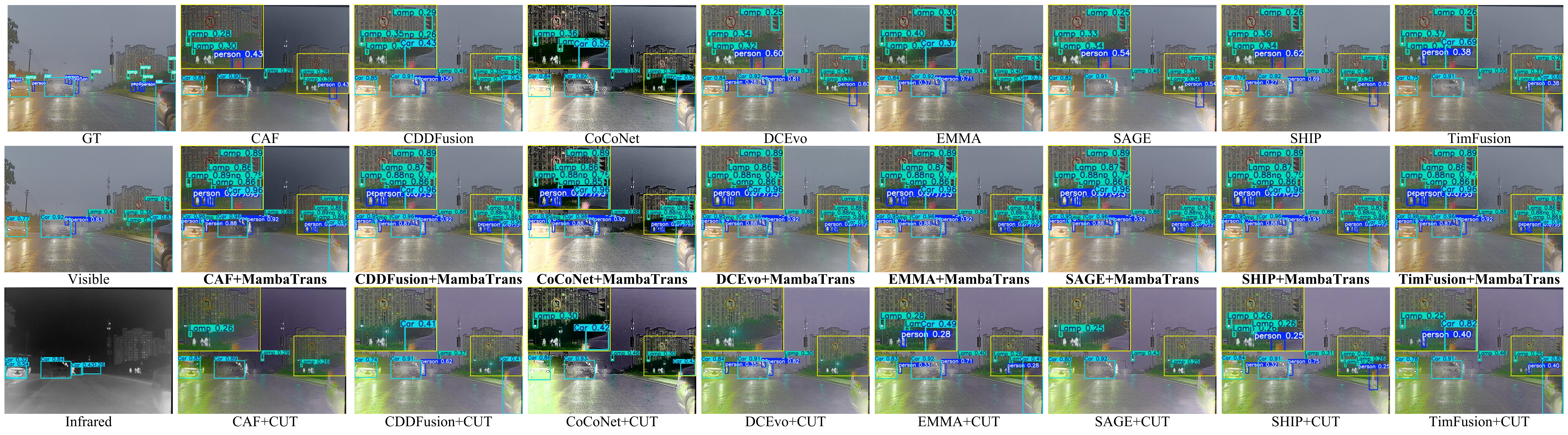}
\caption{Comparison of Deteiction Results of YOLOv11.}
\label{Fig5}
\end{figure}

This section conducts target detection experiments. The results generated by the above 8 fusion methods are translated through the proposed method, and YOLOv11 is used to detect the images. Figure \ref{Fig5} shows that existing methods have difficulty effectively detecting pedestrians beside traffic lights, while the translated images by MambaTrans can achieve effective detection.
Table \ref{tab:23} presents the specific detection metrics. Similar to the segmentation results, the detection performance of visible light images is superior to most of the fused images. MambaTrans can effectively improve the detection performance of fused images, achieving improvements in all categories, and some translation results even outperform the detection metrics of visible light images.

\subsection{Ablation Experiment}

\begin{table}[h!]
\centering
\caption{The influence of different ablation experiments on MM-SSB, with \colorbox[HTML]{C6EFCE}{\color[HTML]{006100}green} for the best.}
\label{tab:4}
\resizebox{\textwidth}{!}{%
\begin{tabular}{cccccccc|ccccccccc}
\hline
             & \multicolumn{7}{c|}{M\textsuperscript{3}FD}                                      & \multicolumn{9}{c}{MSRS}                                                                \\ \hline
             & \textbf{Person} & \textbf{Car}   & \textbf{Bus}   & \textbf{Lamp}  & \textbf{Motorcycle} & \textbf{Truck} & \textbf{mAP50}    & \textbf{Car}   & \textbf{Person} & \textbf{Bike}  & \textbf{Curve} & \textbf{Car Stop} & \textbf{Guardrail} & \textbf{Color Cone} & \textbf{Bump}  & \textbf{mAP50}    \\
w/o 3D-SSM   & 0.804  & 0.911 & 0.914 & 0.810  & 0.765      & 0.827 & 0.838   & 0.815 & 0.645  & 0.509 & 0.702 & 0.617     & 0.892     & 0.691       & 0.811  & 0.710 \\
w/o MM-CA    & 0.868  & 0.919 & 0.928 & 0.819 & 0.755      & 0.840  & 0.854 & 0.882 & 0.690   & 0.544 & 0.720  & 0.675     & 0.890      & 0.683       & 0.836 & 0.740     \\
w/o TAC-Loss & 0.876  & 0.933 & 0.927 & 0.824 & 0.734      & 0.804 & 0.849 & 0.895 & 0.637  & 0.566 & 0.748 & 0.623     & 0.932     & 0.656       & 0.847 & 0.738    \\
Ours &
  \cellcolor[HTML]{C6EFCE}{\color[HTML]{006100} 0.895} &
  \cellcolor[HTML]{C6EFCE}{\color[HTML]{006100} 0.942} &
  \cellcolor[HTML]{C6EFCE}{\color[HTML]{006100} 0.953} &
  \cellcolor[HTML]{C6EFCE}{\color[HTML]{006100} 0.863} &
  \cellcolor[HTML]{C6EFCE}{\color[HTML]{006100} 0.786} &
  \cellcolor[HTML]{C6EFCE}{\color[HTML]{006100} 0.888} &
  \cellcolor[HTML]{C6EFCE}{\color[HTML]{006100} 0.887} &
  \cellcolor[HTML]{C6EFCE}{\color[HTML]{006100} 0.918} &
  \cellcolor[HTML]{C6EFCE}{\color[HTML]{006100} 0.728} &
  \cellcolor[HTML]{C6EFCE}{\color[HTML]{006100} 0.601} &
  \cellcolor[HTML]{C6EFCE}{\color[HTML]{006100} 0.767} &
  \cellcolor[HTML]{C6EFCE}{\color[HTML]{006100} 0.697} &
  \cellcolor[HTML]{C6EFCE}{\color[HTML]{006100} 0.995} &
  \cellcolor[HTML]{C6EFCE}{\color[HTML]{006100} 0.744} &
  \cellcolor[HTML]{C6EFCE}{\color[HTML]{006100} 0.889} &
  \cellcolor[HTML]{C6EFCE}{\color[HTML]{006100} 0.792} \\ \hline
\end{tabular}%
}
\end{table}

MM-SSB, the core component of MambaTrans, enriches the semantic information of Mamba-generated images by integrating text and mask inputs. In this section, we conducted ablation experiments on MM-SSB's components, with results shown in Table \ref{tab:4}. EMMA was used as the fusion method. The findings indicate: (1) Excluding text information in the Selective Scan Module reduces semantic details in translated images, lowering detection and segmentation performance. (2) Without multimodal cross-attention, directly cross-attending text with images for translation fails to form fine textures in target regions due to missing mask information, leading to sub-optimal localization. (3) Using only Charbonnier Loss without detection loss significantly decreases detection performance of translated images due to the lack of object detection priors.

\section{Conclusion}
This paper presents a Mamba-based translator, MambaTrans. This model utilizes the prior knowledge of multimodal large language models and takes the fusion of images, masks, and the text generated by multimodal large language models as input, with the aim of making the distribution of the fused images in the pre-trained downstream task model closer to that of visible images, while enhancing the semantic information within the images and generating fine textures for the main target area for precise localization. Theoretically, this model is adaptable to any existing fusion methods. Experimental results indicate that MambaTrans retains the visual effects of the fused images and exhibits great potential in object detection and semantic segmentation tasks. Furthermore, this model can be extended to the medical field in the future to enhance the segmentation performance of medical images, offering a novel perspective for researchers engaged in multimodal image processing and broader image fusion tasks.

\section*{Acknowledgement}
This research was supported by the National Natural Science Foundation of China (No. 62201149), the Natural Science Foundation of Guangdong Province (No.2024A1515011880), the Basic and Applied Basic Research of Guangdong Province (No.2023A1515140077), the Research Fund of Guangdong-HongKong-Macao Joint Laboratory for Intel-ligent Micro-Nano Optoelectronic Technology (No.2020B1212030010)and the Yunnan Fundamental Research Projects (202301AV070004, 202501AS070123).



\bibliographystyle{unsrtnat}
\bibliography{reference}

\begin{thebibliography}{48}
\providecommand{\natexlab}[1]{#1}
\providecommand{\url}[1]{\texttt{#1}}
\expandafter\ifx\csname urlstyle\endcsname\relax
  \providecommand{\doi}[1]{doi: #1}\else
  \providecommand{\doi}{doi: \begingroup \urlstyle{rm}\Url}\fi

\bibitem[Li et~al.(2024{\natexlab{a}})Li, Liu, Zhang, and Liu]{li2024deep}
Huafeng Li, Junyu Liu, Yafei Zhang, and Yu~Liu.
\newblock A deep learning framework for infrared and visible image fusion without strict registration.
\newblock \emph{International Journal of Computer Vision}, 132\penalty0 (5):\penalty0 1625--1644, 2024{\natexlab{a}}.

\bibitem[Li et~al.(2024{\natexlab{b}})Li, Liu, Li, Zhou, Li, and Nie]{li2024all}
Xilai Li, Wuyang Liu, Xiaosong Li, Fuqiang Zhou, Huafeng Li, and Feiping Nie.
\newblock All-weather multi-modality image fusion: Unified framework and 100k benchmark.
\newblock \emph{arXiv preprint arXiv:2402.02090}, 2024{\natexlab{b}}.

\bibitem[Li et~al.(2023{\natexlab{a}})Li, Tan, Zhou, Wang, and Li]{li2023infrared}
Xilai Li, Haishu Tan, Fuqiang Zhou, Gao Wang, and Xiaosong Li.
\newblock Infrared and visible image fusion based on domain transform filtering and sparse representation.
\newblock \emph{Infrared Physics \& Technology}, 131:\penalty0 104701, 2023{\natexlab{a}}.

\bibitem[Liu et~al.(2024{\natexlab{a}})Liu, Wu, Liu, Wang, Jiang, Ma, Zhong, and Fan]{2}
Jinyuan Liu, Guanyao Wu, Zhu Liu, Di~Wang, Zhiying Jiang, Long Ma, Wei Zhong, and Xin Fan.
\newblock Infrared and visible image fusion: From data compatibility to task adaption.
\newblock \emph{IEEE Transactions on Pattern Analysis and Machine Intelligence}, 2024{\natexlab{a}}.

\bibitem[Xu et~al.(2024)Xu, Li, Jie, and Tan]{xu2024simultaneous}
Yushen Xu, Xiaosong Li, Yuchan Jie, and Haishu Tan.
\newblock Simultaneous tri-modal medical image fusion and super-resolution using conditional diffusion model.
\newblock In \emph{International Conference on Medical Image Computing and Computer-Assisted Intervention}, pages 635--645. Springer, 2024.

\bibitem[Khanam and Hussain(2024{\natexlab{a}})]{3}
Rahima Khanam and Muhammad Hussain.
\newblock Yolov11: An overview of the key architectural enhancements.
\newblock \emph{arXiv preprint arXiv:2410.17725}, 2024{\natexlab{a}}.

\bibitem[Girshick(2015)]{4}
Ross Girshick.
\newblock Fast r-cnn.
\newblock In \emph{Proceedings of the IEEE international conference on computer vision}, pages 1440--1448, 2015.

\bibitem[Lin et~al.(2014)Lin, Maire, Belongie, Hays, Perona, Ramanan, Doll{\'a}r, and Zitnick]{5}
Tsung-Yi Lin, Michael Maire, Serge Belongie, James Hays, Pietro Perona, Deva Ramanan, Piotr Doll{\'a}r, and C~Lawrence Zitnick.
\newblock Microsoft coco: Common objects in context.
\newblock In \emph{Computer vision--ECCV 2014: 13th European conference, zurich, Switzerland, September 6-12, 2014, proceedings, part v 13}, pages 740--755. Springer, 2014.

\bibitem[Van~der Maaten and Hinton(2008)]{6}
Laurens Van~der Maaten and Geoffrey Hinton.
\newblock Visualizing data using t-sne.
\newblock \emph{Journal of machine learning research}, 9\penalty0 (11), 2008.

\bibitem[Jie et~al.(2024)Jie, Xu, Li, and Tan]{7}
Yuchan Jie, Yushen Xu, Xiaosong Li, and Haishu Tan.
\newblock Tsjnet: A multi-modality target and semantic awareness joint-driven image fusion network.
\newblock \emph{arXiv preprint arXiv:2402.01212}, 2024.

\bibitem[Liu et~al.(2022)Liu, Fan, Huang, Wu, Liu, Zhong, and Luo]{8}
Jinyuan Liu, Xin Fan, Zhanbo Huang, Guanyao Wu, Risheng Liu, Wei Zhong, and Zhongxuan Luo.
\newblock Target-aware dual adversarial learning and a multi-scenario multi-modality benchmark to fuse infrared and visible for object detection.
\newblock In \emph{Proceedings of the IEEE/CVF conference on computer vision and pattern recognition}, pages 5802--5811, 2022.

\bibitem[Jie et~al.(2025)Jie, Xu, Li, Zhou, Lv, and Li]{jie2025fs}
Yuchan Jie, Yushen Xu, Xiaosong Li, Fuqiang Zhou, Jianming Lv, and Huafeng Li.
\newblock Fs-diff: Semantic guidance and clarity-aware simultaneous multimodal image fusion and super-resolution.
\newblock \emph{Information Fusion}, 121:\penalty0 103146, 2025.

\bibitem[Liu et~al.(2024{\natexlab{b}})Liu, Liu, Liu, Fan, and Luo]{9}
Risheng Liu, Zhu Liu, Jinyuan Liu, Xin Fan, and Zhongxuan Luo.
\newblock A task-guided, implicitly-searched and metainitialized deep model for image fusion.
\newblock \emph{IEEE Transactions on Pattern Analysis and Machine Intelligence}, 2024{\natexlab{b}}.

\bibitem[Tang et~al.(2022{\natexlab{a}})Tang, Yuan, and Ma]{10}
Linfeng Tang, Jiteng Yuan, and Jiayi Ma.
\newblock Image fusion in the loop of high-level vision tasks: A semantic-aware real-time infrared and visible image fusion network.
\newblock \emph{Information Fusion}, 82:\penalty0 28--42, 2022{\natexlab{a}}.

\bibitem[Zhang et~al.(2024)Zhang, Zuo, Jiang, Guo, and Ma]{11}
Hao Zhang, Xuhui Zuo, Jie Jiang, Chunchao Guo, and Jiayi Ma.
\newblock Mrfs: Mutually reinforcing image fusion and segmentation.
\newblock In \emph{Proceedings of the IEEE/CVF Conference on Computer Vision and Pattern Recognition}, pages 26974--26983, 2024.

\bibitem[Pang et~al.(2021)Pang, Lin, Qin, and Chen]{12}
Yingxue Pang, Jianxin Lin, Tao Qin, and Zhibo Chen.
\newblock Image-to-image translation: Methods and applications.
\newblock \emph{IEEE Transactions on Multimedia}, 24:\penalty0 3859--3881, 2021.

\bibitem[Zhao et~al.(2022)Zhao, Zhong, Luo, Lee, and Sebe]{13}
Yuyang Zhao, Zhun Zhong, Zhiming Luo, Gim~Hee Lee, and Nicu Sebe.
\newblock Source-free open compound domain adaptation in semantic segmentation.
\newblock \emph{IEEE Transactions on Circuits and Systems for Video Technology}, 32\penalty0 (10):\penalty0 7019--7032, 2022.

\bibitem[{\"O}zkano{\u{g}}lu and Ozer(2022)]{14}
Mehmet~Akif {\"O}zkano{\u{g}}lu and Sedat Ozer.
\newblock Infragan: A gan architecture to transfer visible images to infrared domain.
\newblock \emph{Pattern Recognition Letters}, 155:\penalty0 69--76, 2022.

\bibitem[Herrmann et~al.(2018)Herrmann, Ruf, and Beyerer]{15}
Christian Herrmann, Miriam Ruf, and J{\"u}rgen Beyerer.
\newblock Cnn-based thermal infrared person detection by domain adaptation.
\newblock In \emph{Autonomous Systems: Sensors, Vehicles, Security, and the Internet of Everything}, volume 10643, pages 38--43. SPIE, 2018.

\bibitem[Medeiros et~al.(2024{\natexlab{a}})Medeiros, Pena, Aminbeidokhti, Dubail, Granger, and Pedersoli]{16}
Heitor~Rapela Medeiros, Fidel A~Guerrero Pena, Masih Aminbeidokhti, Thomas Dubail, Eric Granger, and Marco Pedersoli.
\newblock Hallucidet: hallucinating rgb modality for person detection through privileged information.
\newblock In \emph{Proceedings of the IEEE/CVF Winter Conference on Applications of Computer Vision}, pages 1444--1453, 2024{\natexlab{a}}.

\bibitem[Medeiros et~al.(2024{\natexlab{b}})Medeiros, Aminbeidokhti, Pe{\~n}a, Latortue, Granger, and Pedersoli]{17}
Heitor~Rapela Medeiros, Masih Aminbeidokhti, Fidel Alejandro~Guerrero Pe{\~n}a, David Latortue, Eric Granger, and Marco Pedersoli.
\newblock Modality translation for object detection adaptation without forgetting prior knowledge.
\newblock In \emph{European Conference on Computer Vision}, pages 51--68. Springer, 2024{\natexlab{b}}.

\bibitem[Gu and Dao(2023)]{18}
Albert Gu and Tri Dao.
\newblock Mamba: Linear-time sequence modeling with selective state spaces.
\newblock \emph{arXiv preprint arXiv:2312.00752}, 2023.

\bibitem[Liu et~al.(2024{\natexlab{c}})Liu, Tian, Zhao, Yu, Xie, Wang, Ye, Jiao, and Liu]{19}
Yue Liu, Yunjie Tian, Yuzhong Zhao, Hongtian Yu, Lingxi Xie, Yaowei Wang, Qixiang Ye, Jianbin Jiao, and Yunfan Liu.
\newblock Vmamba: Visual state space model.
\newblock \emph{Advances in neural information processing systems}, 37:\penalty0 103031--103063, 2024{\natexlab{c}}.

\bibitem[Guo et~al.(2024)Guo, Li, Dai, Ouyang, Ren, and Xia]{20}
Hang Guo, Jinmin Li, Tao Dai, Zhihao Ouyang, Xudong Ren, and Shu-Tao Xia.
\newblock Mambair: A simple baseline for image restoration with state-space model.
\newblock In \emph{European conference on computer vision}, pages 222--241. Springer, 2024.

\bibitem[Zhu et~al.(2024)Zhu, Liao, Zhang, Wang, Liu, and Wang]{21}
Lianghui Zhu, Bencheng Liao, Qian Zhang, Xinlong Wang, Wenyu Liu, and Xinggang Wang.
\newblock Vision mamba: efficient visual representation learning with bidirectional state space model.
\newblock In \emph{Proceedings of the 41st International Conference on Machine Learning}, ICML'24. JMLR.org, 2024.

\bibitem[Shazeer(2020)]{22}
Noam Shazeer.
\newblock Glu variants improve transformer.
\newblock \emph{arXiv preprint arXiv:2002.05202}, 2020.

\bibitem[Ren et~al.(2015)Ren, He, Girshick, and Sun]{ren2015faster}
Shaoqing Ren, Kaiming He, Ross Girshick, and Jian Sun.
\newblock Faster r-cnn: Towards real-time object detection with region proposal networks.
\newblock \emph{Advances in neural information processing systems}, 28, 2015.

\bibitem[Tang et~al.(2022{\natexlab{b}})Tang, Yuan, Zhang, Jiang, and Ma]{23}
Linfeng Tang, Jiteng Yuan, Hao Zhang, Xingyu Jiang, and Jiayi Ma.
\newblock Piafusion: A progressive infrared and visible image fusion network based on illumination aware.
\newblock \emph{Information Fusion}, 83:\penalty0 79--92, 2022{\natexlab{b}}.

\bibitem[Wu et~al.(2025)Wu, Liu, Fu, Peng, Liu, Fan, and Liu]{24}
Guanyao Wu, Haoyu Liu, Hongming Fu, Yichuan Peng, Jinyuan Liu, Xin Fan, and Risheng Liu.
\newblock Every sam drop counts: Embracing semantic priors for multi-modality image fusion and beyond.
\newblock \emph{arXiv preprint arXiv:2503.01210}, 2025.

\bibitem[Liu et~al.(2025)Liu, Zhang, Mei, Li, Zou, Jiang, Ma, Liu, and Fan]{25}
Jinyuan Liu, Bowei Zhang, Qingyun Mei, Xingyuan Li, Yang Zou, Zhiying Jiang, Long Ma, Risheng Liu, and Xin Fan.
\newblock Dcevo: Discriminative cross-dimensional evolutionary learning for infrared and visible image fusion.
\newblock \emph{arXiv preprint arXiv:2503.17673}, 2025.

\bibitem[Liu et~al.(2024{\natexlab{d}})Liu, Wu, Liu, Ma, Liu, and Fan]{26}
Jinyuan Liu, Guanyao Wu, Zhu Liu, Long Ma, Risheng Liu, and Xin Fan.
\newblock Where elegance meets precision: towards a compact, automatic, and flexible framework for multi-modality image fusion and applications.
\newblock In \emph{Proceedings of the Thirty-Third International Joint Conference on Artificial Intelligence}, pages 1110--1118, 2024{\natexlab{d}}.

\bibitem[Zhao et~al.(2024)Zhao, Bai, Zhang, Zhang, Zhang, Xu, Chen, Timofte, and Van~Gool]{28}
Zixiang Zhao, Haowen Bai, Jiangshe Zhang, Yulun Zhang, Kai Zhang, Shuang Xu, Dongdong Chen, Radu Timofte, and Luc Van~Gool.
\newblock Equivariant multi-modality image fusion.
\newblock In \emph{Proceedings of the IEEE/CVF conference on computer vision and pattern recognition}, pages 25912--25921, 2024.

\bibitem[Zhao et~al.(2023)Zhao, Bai, Zhang, Zhang, Xu, Lin, Timofte, and Van~Gool]{29}
Zixiang Zhao, Haowen Bai, Jiangshe Zhang, Yulun Zhang, Shuang Xu, Zudi Lin, Radu Timofte, and Luc Van~Gool.
\newblock Cddfuse: Correlation-driven dual-branch feature decomposition for multi-modality image fusion.
\newblock In \emph{Proceedings of the IEEE/CVF conference on computer vision and pattern recognition}, pages 5906--5916, 2023.

\bibitem[Zheng et~al.(2024)Zheng, Zhou, Huang, Hou, Li, Xu, and Zhao]{30}
Naishan Zheng, Man Zhou, Jie Huang, Junming Hou, Haoying Li, Yuan Xu, and Feng Zhao.
\newblock Probing synergistic high-order interaction in infrared and visible image fusion.
\newblock In \emph{Proceedings of the IEEE/CVF conference on computer vision and pattern recognition}, pages 26384--26395, 2024.

\bibitem[Liu et~al.(2024{\natexlab{e}})Liu, Lin, Wu, Liu, Luo, and Fan]{31}
Jinyuan Liu, Runjia Lin, Guanyao Wu, Risheng Liu, Zhongxuan Luo, and Xin Fan.
\newblock Coconet: Coupled contrastive learning network with multi-level feature ensemble for multi-modality image fusion.
\newblock \emph{International Journal of Computer Vision}, 132\penalty0 (5):\penalty0 1748--1775, 2024{\natexlab{e}}.

\bibitem[Park et~al.(2020{\natexlab{a}})Park, Efros, Zhang, and Zhu]{33}
Taesung Park, Alexei~A Efros, Richard Zhang, and Jun-Yan Zhu.
\newblock Contrastive learning for unpaired image-to-image translation.
\newblock In \emph{Computer Vision--ECCV 2020: 16th European Conference, Glasgow, UK, August 23--28, 2020, Proceedings, Part IX 16}, pages 319--345. Springer, 2020{\natexlab{a}}.

\bibitem[Everingham et~al.(2015)Everingham, Eslami, Van~Gool, Williams, Winn, and Zisserman]{34}
Mark Everingham, SM~Ali Eslami, Luc Van~Gool, Christopher~KI Williams, John Winn, and Andrew Zisserman.
\newblock The pascal visual object classes challenge: A retrospective.
\newblock \emph{International journal of computer vision}, 111:\penalty0 98--136, 2015.

\bibitem[Roberts et~al.(2008)Roberts, Van~Aardt, and Ahmed]{35}
J~Wesley Roberts, Jan~A Van~Aardt, and Fethi~Babikker Ahmed.
\newblock Assessment of image fusion procedures using entropy, image quality, and multispectral classification.
\newblock \emph{Journal of Applied Remote Sensing}, 2\penalty0 (1):\penalty0 023522, 2008.

\bibitem[Cui et~al.(2015)Cui, Feng, Xu, Li, and Chen]{36}
Guangmang Cui, Huajun Feng, Zhihai Xu, Qi~Li, and Yueting Chen.
\newblock Detail preserved fusion of visible and infrared images using regional saliency extraction and multi-scale image decomposition.
\newblock \emph{Optics Communications}, 341:\penalty0 199--209, 2015.

\bibitem[Zhang et~al.(2011)Zhang, Zhang, Mou, and Zhang]{37}
Lin Zhang, Lei Zhang, Xuanqin Mou, and David Zhang.
\newblock Fsim: A feature similarity index for image quality assessment.
\newblock \emph{IEEE transactions on Image Processing}, 20\penalty0 (8):\penalty0 2378--2386, 2011.

\bibitem[Cvejic et~al.(2006)Cvejic, Canagarajah, and Bull]{38}
N~Cvejic, CN~Canagarajah, and DR~Bull.
\newblock Image fusion metric based on mutual information and tsallis entropy.
\newblock \emph{Electronics letters}, 42\penalty0 (11):\penalty0 626--627, 2006.

\bibitem[Eskicioglu and Fisher(2002)]{39}
Ahmet~M Eskicioglu and Paul~S Fisher.
\newblock Image quality measures and their performance.
\newblock \emph{IEEE Transactions on communications}, 43\penalty0 (12):\penalty0 2959--2965, 2002.

\bibitem[Park et~al.(2020{\natexlab{b}})Park, Efros, Zhang, and Zhu]{40}
Taesung Park, Alexei~A Efros, Richard Zhang, and Jun-Yan Zhu.
\newblock Contrastive learning for unpaired image-to-image translation.
\newblock In \emph{Computer Vision--ECCV 2020: 16th European Conference, Glasgow, UK, August 23--28, 2020, Proceedings, Part IX 16}, pages 319--345. Springer, 2020{\natexlab{b}}.

\bibitem[Doubao()]{volcengine-doubao-vision-pro-32k}
Doubao.
\newblock Doubao vision pro 32k.
\newblock \url{https://console.volcengine.com/ark/region:ark+cn-beijing/model/detail?Id=doubao-vision-pro-32k}.

\bibitem[He et~al.(2023)He, Li, Geng, and Xie]{he2023fastinst}
Junjie He, Pengyu Li, Yifeng Geng, and Xuansong Xie.
\newblock Fastinst: A simple query-based model for real-time instance segmentation.
\newblock In \emph{Proceedings of the IEEE/CVF conference on computer vision and pattern recognition}, pages 23663--23672, 2023.

\bibitem[Cai and Vasconcelos(2018)]{cai2018cascade}
Zhaowei Cai and Nuno Vasconcelos.
\newblock Cascade r-cnn: Delving into high quality object detection.
\newblock In \emph{Proceedings of the IEEE conference on computer vision and pattern recognition}, pages 6154--6162, 2018.

\bibitem[Li et~al.(2023{\natexlab{b}})Li, Zhang, Xu, Liu, Zhang, Ni, and Shum]{li2023mask}
Feng Li, Hao Zhang, Huaizhe Xu, Shilong Liu, Lei Zhang, Lionel~M Ni, and Heung-Yeung Shum.
\newblock Mask dino: Towards a unified transformer-based framework for object detection and segmentation.
\newblock In \emph{Proceedings of the IEEE/CVF conference on computer vision and pattern recognition}, pages 3041--3050, 2023{\natexlab{b}}.

\bibitem[Khanam and Hussain(2024{\natexlab{b}})]{khanam2024yolov11}
Rahima Khanam and Muhammad Hussain.
\newblock Yolov11: An overview of the key architectural enhancements.
\newblock \emph{arXiv preprint arXiv:2410.17725}, 2024{\natexlab{b}}.

\end{thebibliography}







\appendix

\end{document}